# BIOLOGICALLY INSPIRED DESIGN CONCEPT GENERATION USING GENERATIVE PRE-TRAINED TRANSFORMERS

**Qihao Zhu**
Data-Driven Innovation Lab
Singapore University of Technology and Design
qihao_zhu@mymail.sutd.edu.sg

**Xinyu Zhang**
State Key Laboratory of Automotive Safety
and Energy
Tsinghua University
xyzhang@tsinghua.edu.cn

**Jianxi Luo**
Data-Driven Innovation Lab
Singapore University of Technology and Design
jianxi_luo@sutd.edu.sg

December 26, 2022

## ABSTRACT

Biological systems in nature have evolved for millions of years to adapt and survive the environment. Many features they developed can be inspirational and beneficial for solving technical problems in modern industries. This leads to a specific form of design-by-analogy called bio-inspired design (BID). Although BID as a design method has been proven beneficial, the gap between biology and engineering continuously hinders designers from effectively applying the method. Therefore, we explore the recent advance of artificial intelligence (AI) for a data-driven approach to bridge the gap. This paper proposes a generative design approach based on the generative pre-trained language model (PLM) to automatically retrieve and map biological analogy and generate BID in the form of natural language. The latest generative pre-trained transformer, namely GPT-3, is used as the base PLM. Three types of design concept generators are identified and fine-tuned from the PLM according to the looseness of the problem space representation. Machine evaluators are also fine-tuned to assess the mapping relevancy between the domains within the generated BID concepts. The approach is evaluated and then employed in a real-world project of designing light-weighted flying cars during its conceptual design phase The results show our approach can generate BID concepts with good performance.

## 1 Introduction

Bio-inspired design (BID) is a design methodology in the engineering design field that is based on the observation of biological natural phenomena [1] and aims to develop novel solutions for real-world problems with analogies from nature [2]. For millions of years, creatures on earth managed to evolve features to survive the environments and compete with other species, and some of them could be crucial for solving the technical challenges we are facing today. In a high-speed train design example [3], engineers observed that kingfishers are capable of diving into the water from air with extremely high speed without making a splash and took the kingfisher's beak shape as an analogy to develop the head of the high-speed train. This improvement not only removed the sonic boom caused by the train traveling at high speed through a tunnel, but also resulted in a faster speed and lower energy consumption.

Although the BID method has been proven helpful and become popular among designers and researchers, it is challenging for designers to find, recognize, and understand biological information in design practice because they lack the necessary knowledge and experience in the biology field [4, 5]. BID successes are often serendipitous. For such challenges related to knowledge discovery and information processing, a data-driven approach and knowledge-based artificial intelligence (AI) can aid human designers for BID [6]. Recent advances in deep learning show particular promises if we can properly apply them to biological data and design problems and bridge the gap. Our goal is to employ AI to assist the retrieval and mapping of biological analogies and the generation of inspirational BID concepts, helping designers to explore diverse and novel analogical opportunities that they could hardly think of without extensive reserves of biological knowledge.

In this paper, we apply one of the latest natural language processing (NLP) techniques to help designers innovate with BID. We propose a natural language generation approach using the pre-trained language model (PLM), more specifically, the generative pre-trained transformer (GPT), to automate the retrieval and mapping of biological information through text generation. The generated concepts can be evaluated through classifiers fine-tuned from the same base PLM. This could generate and evaluate many BID concepts in a very short period, and the natural language representation of the results are understandable for designers. We further employed the approach in a real flying car design project and tested its performance to solve a real-world technical problem together with a team of flying car experts.

## 2 Background and Related Works

### 2.1 Design-by-Analogy

Design-by-analogy (DbA) is a design methodology that innovates the design target through the analogy with a different domain [7], which is proven effective to help overcome design fixation [8]. Analogy in design is generally defined as "illustration of an idea by means of another familiar idea that is similar or parallel to it in some significant features" [9]. This indicates two underlying domains in analogy reasoning. The domain that contains the problem and needs to be understood is called the target domain, while the one that provides a potential solution to the problem is called the source domain [7, 9, 10].

Various prior studies have investigated the cognitive process and reasoning behind DbA, and they basically agree on three important subphases: retrieval, mapping, and evaluation [9-12]. The retrieval of source for analogy depends on the relational similarity between domains [10, 13, 14]. This means that the retrieved source domain should share a common relationship with the target domain regarding how they are comprised by their components. Then, mapping between the target and the retrieved source is established and evaluated. This is usually done by comparing the elements and patterns between domains [15, 16].

In recent years, many data-driven methods and tools for design-by-analogy have been proposed. For example, [17, 18] unify systems and methods in diverse engineering disciplines to support transferring knowledge between different domains. Others develop knowledge graph and semantic network based on patents data to guide the exploration and retrieval of analogies in the vast technology space [19-21]. Readers may refer to [6] for the latest and comprehensive review of data-driven design-by-analogy research. In this study, we focus on bio-inspired design (BID), which is a special form of DbA, and explore a pre-trained language model to automate the retrieval, mapping, and evaluation of biological analogies for design concept generation.

### 2.2 Bio-Inspired Design

Bio-inspired design (BID) is an innovation method whereby designers take inspiration from biological phenomena. Although some researchers argue that BID goes beyond analogy [22], it is commonly interpreted as a subset of DbA where the source domain is specified as biological systems from nature [1, 23-25]. The methodology has grown into an innovative design process in the field of engineering design which leverages analogies to biological systems to develop varied and novel solutions for engineering challenges [2, 25].

According to literatures, there are two main approaches for bio-inspired design: problem-driven and solution-based [26, 27]. The problem-driven approach is a biomimetic development method that aims to address a practical problem, with the problem serving as the process's starting point. By contrast, the solution-based approach seeks inspiration from the observation of nature and allows the identification of a design problem, and therefore the knowledge of a biological system of interest serves as the basis for design. For this paper, we focus on the problem-driven approach as it better fits the general design process model proposed by [28].

Researchers have identified three domains in problem-driven BID: problem domain, nature domain, and solution domain [27]. Generally, the problem domain and solution domain represent the situations of the target domain before and after analogy mapping, while the nature domain is corresponding to the source domain which indicates the biological features used as analogy source. However, it is universally acknowledged that the transitions between the three domains are challenging due to the distinctions between engineering and biology [5, 22, 29]. Therefore, many contributions in recent years have explored methods and tools trying to bridge gaps among the three domains in BID.

The transition from problem domain to nature domain includes the abstraction of the engineering problem and the identification of the biological system [29, 30]. For abstraction, [31] employ the functional model of the desired engineering system for representing the essence of functionality, which is then used to explore biological solutions for inspiration. [32] propose a natural language processing-based process model that begins with the formulation of original functional terms to represent a problem. For biological system identification, AskNature [33] groups biological strategies and innovations by function according to the Biomimicry Taxonomy, so that it encodes biological information in engineering terms. Another popular attempt on this issue is to search or push biology keywords based on technical problems. Shu & Cheong [34] explore a natural language approach to search for biology keywords for analogy. Chen, Li, et al. [35] also proposed a method to automatically push keywords based on Composite Correlation Intension values.

The gap between nature domain and solution domain is believed to be the result of the challenge of designers' understandability [5]. And, the representations of biological systems are key to knowledge transfer in BID process [36]. To bridge this gap, AskNature [33] provides biomimicry strategies that help designers develop a deeper understanding of how to apply the information within the nature domain. Sartori, et al. [36] developed the SAPPhIRE (State-changes, Actions, Parts, Phenomena, Inputs, oRgans and Effects) guideline to formulate four distinct levels of abstraction at which the transfer between domains could take place. Chen, Tao, et al. [37] proposed a method combining dependency parsing and keyword extraction to extract structure-function knowledge from nature domain.

## 3 Research Method

### 3.1 Pre-Trained Language Model (PLM)

To bridge the gaps among the three domains (i.e., nature, solution, and problem) in BID, we propose a natural language generation approach to create understandable textual concepts. This is accomplished by the generation and evaluation of design concepts through the application of the pre-trained language model (PLM).

PLMs are language models that have been trained with a large dataset of textual information and can be applied to deal with various language-related tasks [38]. For example, the base model of BERT was trained with Wiki and books data that contains over 3.3 billion tokens [39], and the largest PLM nowadays, namely GPT-3, was trained on a 400 billion tokens dataset [40]. These huge pre-training datasets offer PLMs not only the capability of understanding human language, but also the knowledge and logic that come with it.

PLMs such as BERT [39] or GPT [40, 41] have been increasingly popular in the natural language processing (NLP) field for their state-of-the-art performance in many downstream NLP tasks [42] including text completion and text classification. There are mainly two mechanisms for a PLM to perform these NLP tasks: fine-tuning and prompt-based learning. Fine-tuning, first proposed by [43], is a technique to re-train the pre-trained model with a small amount of task-specific dataset for the task of interest. The parameters of the base model are updated in the process. Prompt-based learning, on the other hand, makes no change to the base model but can leverage what the model has already learned with simple prompts. This mechanism leads to the few-shot learning technique which requires only a few examples of the desired task as prompt [40].

In this paper, the PLM we use for concept generation and evaluation is the latest version of generative pre-trained transformer (GPT), i.e., GPT-3 [40], and we fine-tune the base model for different concept generation tasks in our approach. The reasons that we choose to fine-tune GPT-3 over other models or techniques are as follow:

1. The base model of GPT-3 was pre-trained on much larger dataset than any other PLMs, which makes it more knowledgeable about biology and nature.
2. Fine-tuning can better leverage the BID-related knowledge and logic from the prepared dataset as we assume that previous knowledge of BID is not included in the base PLMs.
3. A large dataset is required to fine-tune a BERT or GPT-2 model which is unobtainable for the BID task, while according to OpenAI [44], GPT-3 only needs a few hundred of high-quality samples for fine-tuning.
4. Masked language models like BERT are generally weak at natural language generation because they can only learn contextual representation of words, namely natural language understanding, but not organize and generate language [42], which makes them unsuitable for our generative approach.

### 3.2 Dataset

The dataset we use to fine-tune the base GPT-3 model is collected from the innovation section of the AskNature website [33] in January 2022. The dataset contains 221 successful BID samples. Each sample is represented by textual information of a benefits, applications, the challenge, innovation details, and a biomimicry story (Table 1).

Table 1: Data composition of each innovation sample in the fine-tuning dataset

| Component | Type of Data | Description |
| --- | --- | --- |
| Benefits | Keywords | The advantages of the innovation. |
| Applications | Keywords | The applications of the innovation. |
| The Challenge | Paragraph | A statement of problem in the target domain that is challenging to solve. |
| Innovation Details | Paragraph | Introduction of the innovation that aims to solve the challenge. |
| Biomimicry Story | Paragraph | A biological natural phenomenon that is applicable as analogy for solving the challenge. |

The variety of information in this dataset offers great opportunities to explore different fine-tuning strategies, and this can be done simply through reformulating and customizing the textual data into input-output pairs. For example, given the input-output pair of applications to innovation details, after fine-tuning the model can generate new innovations based on any applications of user's interest. If we train it the other way around, i.e., using innovation details as input and applications as output, the model can do a very different task that extracts keywords of the applications from any input innovation description. By customizing the AskNature dataset and combining different fine-tuned models, different BID related tasks can be implemented.

## 3.3 Natural Language Processing

Based on both DbA and BID literatures, we propose a methodological framework (Figure 1) for BID concept generation and evaluation through natural language processing. The generation process starts by retrieving a biological system from nature domain based on a given problem space, and the extracted biology information is then mapped to the solution domain. This is done by fine-tuning the GPT-3 model for natural language generation (NLG) with the data sequence of problem, nature, and solution. Sequence is very important for GPTs because they are autoregressive models that predict the next token based on all tokens before. The evaluation process assesses if the mappings between domains have been strongly established during the generation, i.e., if the generated solution aims to solve the given problem, and if it has taken inspiration from the retrieved biological information. These can be analysed through natural language understanding (NLU). Specifically, separate text classifiers can be fine-tuned from the GPT-3 base model to assess the relevancy between problem domain and solution domain and between nature domain and solution domain.

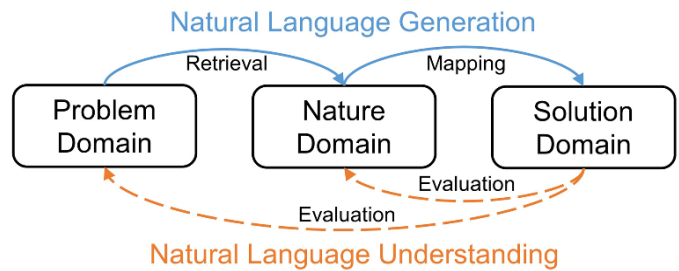

Figure 1: Methodological framework

### 3.3.1 Concept Generators

For concept generation, we identify three types based on the looseness of the problem space (Table 2). Type-1 generation has the minimum constraints regarding how the problem is defined, it takes only the applications as input and generate biomimicry and innovation based on it. This means that the retrieval and mapping of the biological information only take the target's applications into consideration and leaves the problem space wide-open. This type is suitable for scenarios where companies want to innovate a certain product of interest, i.e., drone, but without a specific problem in mind. By hypothesis, Type-1 generator should be able to provide the most diverse results.

Type-2 constrains the problem space by introducing the desired benefits from the potential solutions, i.e., what advantages is this innovation expected to bring. It can be seen as an abstraction of a complex problem into a few keywords of benefits and applications. This type using the abstract representation of the problem is suitable when designers have only a general requirement based on the user's needs, e.g., a small and light-weighted drone.

Type-3 generation further constrains the problem space by using a paragraph of challenge statement, which could contain three to four sentences that describes a problem from the engineering perspective. As the strictest way of problem definition that we can use based on the AskNature dataset, this type could be useful for the engineering teams that may come across a specific and detailed problem in design practice and wish to solve it by BID.

Table 2: Fine-tuned models for generation

| Fine-tuned Model | NLP Task | Input | Output | Description |
|---|---|---|---|---|
| Type-1 Generation (Application) | Generation | Applications | Biomimicry, Innovation | Generate concept with an open problem space |
| Type-2 Generation (Benefits & Application) | Generation | Benefits, Applications | Biomimicry, Innovation | Represent the problem in simple keywords of applications and benefits |
| Type-3 Generation (Challenge) | Generation | Challenge statement | Biomimicry, Innovation | Use the full challenge statement as problem space |

### 3.3.2 Concept Evaluators

Due to the varied problem representation of the three types of generation, different sets of classifiers are required to evaluate the mappings between domains. This includes problem-solution relevancy which uses either benefits and application (Type-2) or challenge statement (Type-3) to represent the problem domain, and nature-solution relevancy (Table 3). However, directly learning relations of logic between texts is difficult for today's NLP techniques. Therefore, we transform these relation learning tasks into simpler binary text classification tasks to predict if the two domains in the given text information are relevant. To construct the dataset for fine-tuning a classifier, we need negative samples that do not acquire the mappings of interest but maintain the form and topic that are as close to the positive samples as possible.

Table 3: Fine-tuned models for evaluation

| Fine-tuned Model | NLP Task | Input | Output | Description |
|---|---|---|---|---|
| Problem-Solution Relevancy Evaluation for Type-2 | Classification | Benefits, Innovation | Related/ Unrelated | Evaluate if the generated innovation may bring the given benefits, applicable to type-2 generation |
| Problem-Solution Relevancy Evaluation for Type-3 | Classification | Challenge, Innovation | Related/ Unrelated | Evaluate if the generated innovation aims to solve the given challenge, applicable to type-3 generation |
| Nature-Solution Relevancy Evaluation | Classification | Biomimicry, Innovation | Related/ Unrelated | Evaluate if the generated innovation takes inspiration from the biomimicry, applicable for all three types of generation |
| Random Innovation Generation | Generation | Application | Innovation | Generate random innovation that is only relevant to the input application. Used to create irrelevant solution to nature or problem domain |

Figure 2 depicts our method to construct the datasets and fine-tune the relevancy evaluators, with the example of evaluating the relevancy between solution (i.e., innovation) and nature (i.e., biomimicry) domains. The two domains that form the positive samples are from the collected AskNature dataset. On the other hand, in negative ones we replace only the innovation into random samples to create the irrelevance. To obtain such negative samples, an additional generator is fine-tuned using application and innovation data (Table 3), which generates random innovations based on only application input (on the right side of Figure 2). In this way, the generated negative sample shares the same application with the positive one (i.e., they design for the same target domain), but without considering the problem and biological inspiration.

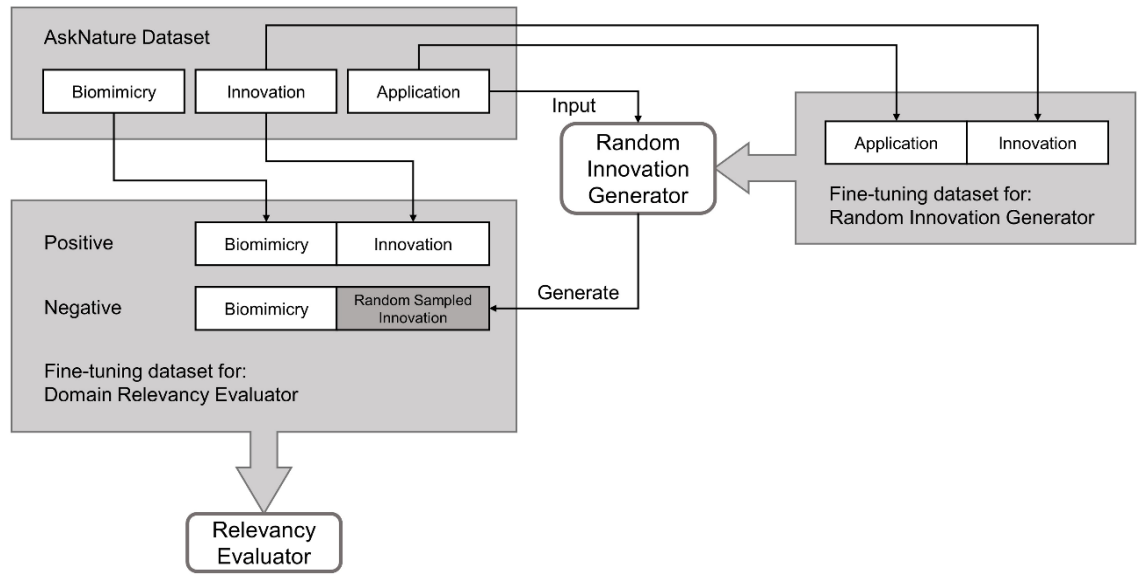

Figure 2: Example workflow for preparing dataset and fine-tuning the domain relevancy evaluators

The fine-tuning datasets for evaluating the relevancy between solution and problem domains can be prepared in the same way. The only difference is that, when evaluating the relevancy of benefits and innovation, we use the sequence of innovation and then benefits. Because it is more natural for human to understand benefits as the result of the innovation, and this reasoning works the same way for GPT-3 during natural language understanding (i.e., assess if the given innovation can result in the desired benefits).

Furthermore, following the suggestion of [45] for a language model to learn relations, marker tokens are added with all samples for fine-tuning the classifiers. Marker tokens are simple tags that are added before and after the sentences or paragraphs of interest. An example of how the marker tokens are used is: *"[Bio]Octopus tentacles have suckers that allow the organisms to hold small objects. The suckers have small…[/Bio][Inno]The soft manipulator was inspired by the suckers of octopus tentacles. It is made of a temperature-responsive layer of…[/Inno]"*. The tags that come with square brackets in the above example are marker tokens.

## 4 Assessing The Concept Generators and Evaluators

### 4.1 Fine-Tuned Generators

The three types of generators (Table 2) are fine-tuned from GPT-3's base model of Davinci, which is currently the largest tunable base model that OpenAI provides via their API. The model was fine-tuned for 4 epochs with the batch size of 1 (batch size is determined based on 0.2% of the size of training samples). The gradually decreasing training loss of all three types illustrated in Figure 3 shows the model's capability of learning BID from our customized training dataset.

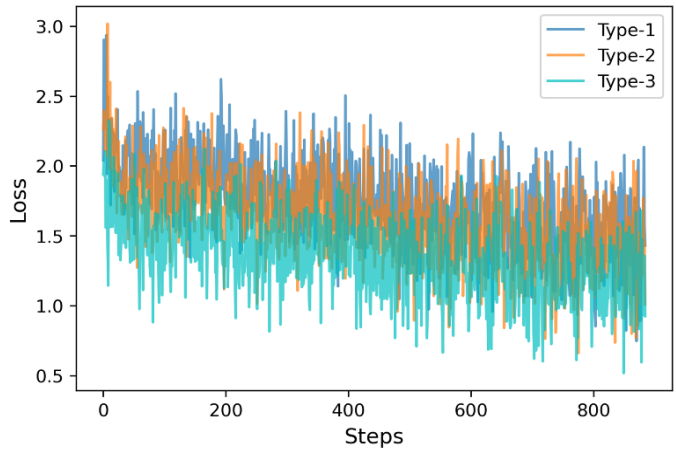

Figure 3: Training loss for fine-tuning three types of concept generation

To assess the fine-tuned models regarding their capability of generating novel design concepts, we employ the metrics of Word Mover's Distance (WMD) [46]. WMD is widely applied in NLP field to measure the semantic distance between two text documents based on word embeddings and is popular in natural language generation (NLG) tasks like image captioning [47] and machine translation [48] to assess the performance of the NLG model. Such applications often aim for minimizing WMD as they want the generated results to be as close to the ground truth as possible. However, we want to do the opposite in our design concept generation task, i.e., assess which model gives the higher WMD. In the concept generation scenario, higher WMD between the generated concepts and the well-established designs represents the concepts are more diverse from existing innovations, and the diversity of concepts could lead to novel solutions. Therefore, we use the metrics to represent the model's capability of generating diverse and potentially novel concepts. Note that this approach can only assess the performance of the generator model but not the novelty of any individual concept generated with customized input, because the well-established ground truth innovation for every new set of problem input can hardly be obtained. In the context of this experiment, the AskNature dataset will provide the existing innovations for comparison.

To measure the WMD, we select three samples of robotics from the AskNature innovation dataset: an aerial robot[1], a ground robot[2], and an underwater robot[3]. Their applications, benefits, and challenges are used as input according to the three types of generation, and 50 concepts are generated for each sample through each type of generator. Then, we extract the word embeddings of the innovation of each original sample and generated ones, measure their semantic distance using WMD. The word embedding extraction method we use in this experiment is the pre-trained Word2Vec model provided by Gensim[4] which was trained on a non-technical dataset of Google News. The word embeddings extracted from this model better represent people's common-sense understanding about the design concepts. In addition, a comparison group is added aiming to provide an approximate baseline of WMD that represents a high diversity in this context. This is done by employing the 36 innovations in the robotics sector of AskNature, measuring the WMD between each of the three samples and all the 35 innovations of robotics (excluding the sample to be measured itself). In this case, the diversity of the AskNature dataset itself is introduced as the baseline because it consists of different applications aiming to solve different problems in different ways.

---

[1] https://asknature.org/innovation/highly-maneuverable-drone-inspired-by-swifts/
[2] https://asknature.org/innovation/speedy-soft-robot-inspired-by-cheetahs/
[3] https://asknature.org/innovation/stretchable-robotic-architecture-inspired-by-pelican-eels/
[4] https://radimrehurek.com/gensim/

Figure 4 (a) (b) (c) shows the WMD distribution of the 50 results of each sample based on generation types. Figure 4 (d) summarizes all the 450 results based on the three types of generation. Although the results differ by the samples, the general trend is that Type-1 generation results in the most diverse concepts that is very close to the baseline diversity of the comparison group, while Type-3 gives the least diverse ones. Type-1 and Type-3 have the loosest and strictest defined problem space as constraints, respectively.

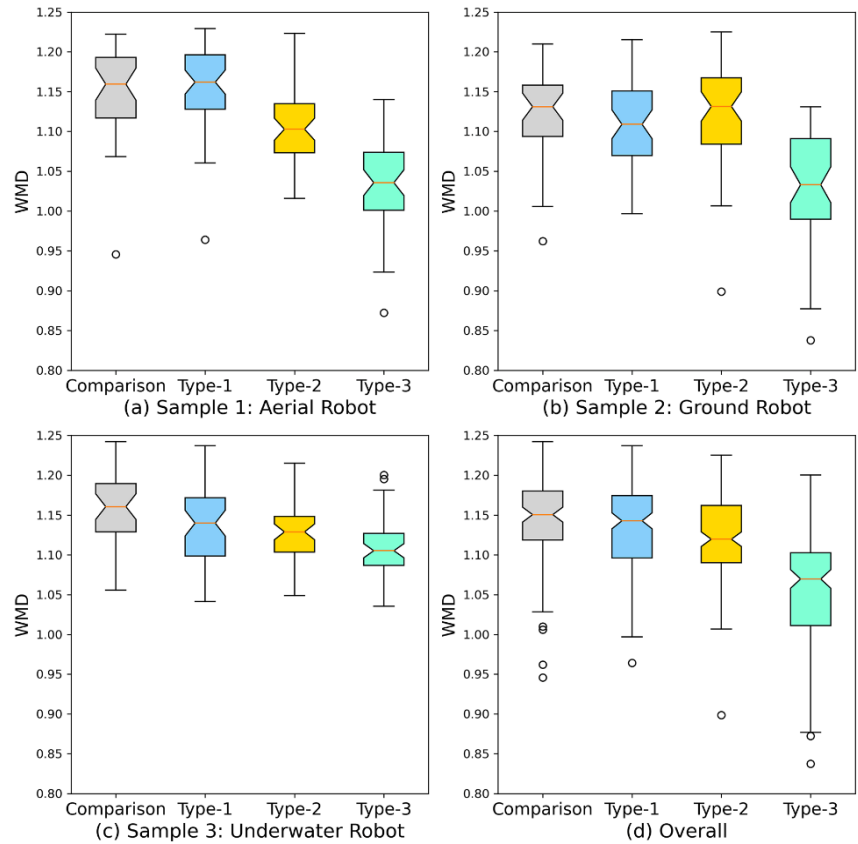

Figure 4: WMD distribution of generated concepts

### 4.2 Fine-Tuned Evaluators

After preparing the datasets for fine-tuning the domains relevancy evaluators following Figure 2, each fine-tuning dataset has 442 samples in total, including 221 positive ones from AskNature and 221 negative ones whose innovations were randomly sampled based on applications. The datasets are then divided into training and validation sets based on the ratio of 4:1.

The evaluators are fine-tuned from GPT-3's base model of Curie using the prepared training set. The model of Curie is not as strong as Davinci but quite capable of complex text classification according to OpenAI [44]. Figure 5 shows the classification accuracy when fine-tuning the evaluators. The accuracy values are returned from OpenAI API after each epoch when fine-tuning the models. The accuracy values are calculated on the validation set. The accuracy of the benefits-innovation evaluator and challenge-innovation evaluator reach 75.3% and 79.8% respectively after 4 epochs, while the biomimicry-innovation evaluator achieves 89.9% accuracy after only 1 epoch, and it further rises to 94.4% accuracy after 2 epochs.

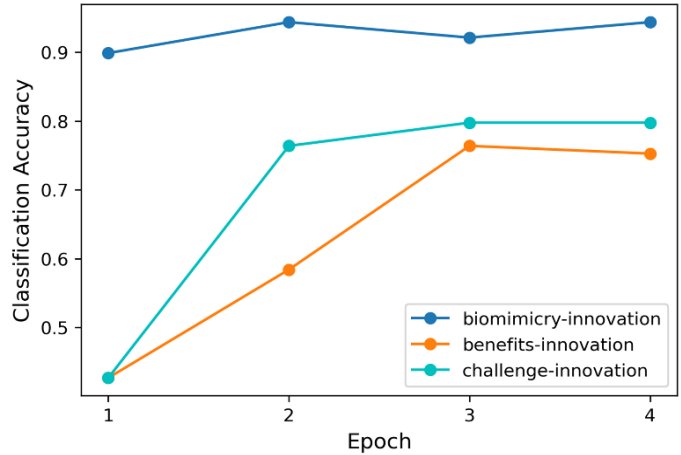

Figure 5: Classification accuracy of the three relevancy evaluators

## 5 Case Study

### 5.1 Project Background

We have applied our BID concept generation approach in a real project to design flying cars. Along with the acceleration of the urbanization process, ground traffic in cities is increasingly crowded. However, the airspace above the urban area has not yet been utilized for transportation. Air transportation provides a prospective solution for road congestion. To merge the air and ground traffic, the flying car is proposed as a class of novel dual-mode vehicles. Figure 6 illustrates three well-known flying car prototypes. However, many technical challenges remain to be solved before we can finally see them flexibly flying and driving in the cities [49].

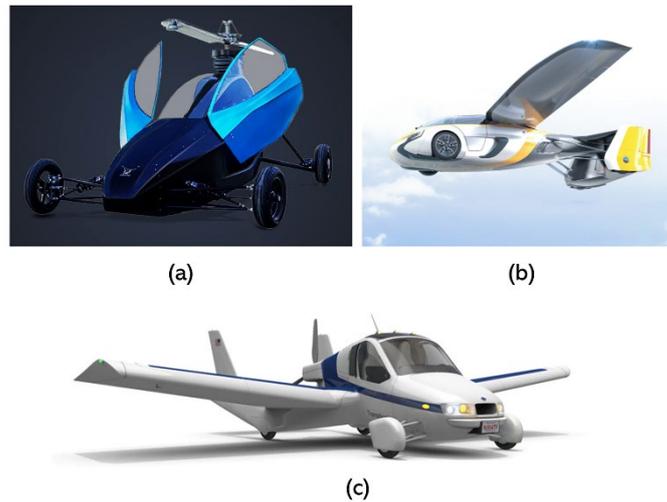

Figure 6: Flying car examples: (a) Pegasus E-model[5]  (b) Aeromobil[6]  (c) Terrafugia Transition[7]

---

[5] https://bepegasus.com/index.jsp
[6] https://www.aeromobil.com/
[7] https://terrafugia.com/transition/

At the beginning of the project, we interviewed the project team of 10 engineers that had worked on several flying car design projects regarding the challenges they came across in their design projects including the present one. Lightweight design emerges as a major design problem of flying cars and requires novel solutions. This is because existing flying car designs normally combine two heavy powertrains for ground and aerial modalities respectively. Table 4 lists the corresponding inputs to all three types of fine-tuned GPT-3 generators. With Type-1, the fine-tuned model is expected generates open-ended flying car concepts. With Type-2 and Type-3, the model is expected to generate lightweight flying car concepts. For each type, 50 concepts are generated.

Table 4: Inputs of flying car case study

| Applications | Flying car |
|---|---|
| Benefits | Lightweight |
| Challenge Statement (Provided by the engineering team) | A flying car includes a subsystem for flying in the air in addition to a subsystem for driving on the ground. With both flying and driving subsystems in one, the weight might be increased to increase the drive load, demand more propulsion to overcome gravity, and increase fuel consumption. Lightweight design is a challenge for flying cars. |

**5.2 Results: Design Concept Generation and Evaluation**

Figure 7 categorizes the biological sources of the GPT-generated bio-inspired flying car design concepts. Across all three types, the generators are over 50% likely to retrieve birds as an analogy, and then insects. This is consistent with the cognitive process of humans where people tend to retrieve information based on attribute similarities [10], i.e., retrieve flying animals as analogies for flying cars. However, there are some unexpected but interesting creatures like reptiles and even plants, which could result in more novel concepts.

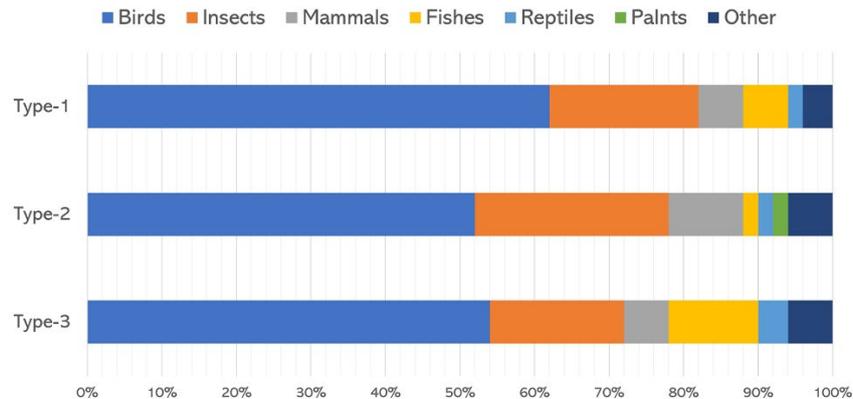

Figure 7: Categories of the retrieved biological systems

The generated concepts are then assessed through the evaluators. Figure 8 shows the probability distribution of the domain relevancy of the generated flying car concepts. Type-1 is excluded for the evaluation of problem-solution relevancy because its problem space is open. Probability represents how confident the classifier is to decide the input domains are relevant. If the probability is lower than 0.5, then the input domains are classified as unrelated. For nature-solution relevancy, the distribution is polarized which shows the evaluator is very confident that most of the concepts do take inspirations from the retrieved biological systems. On the other hand, the probability distributions of problem-solution relevancy appear differently and have peaks that are located above 0.5. This could mean that the patterns to identify if a solution aims to solve the given problem are more difficult to learn for GPT-3. The results of the percentage of generated concepts passing the relevancy evaluations are shown in Table 5.

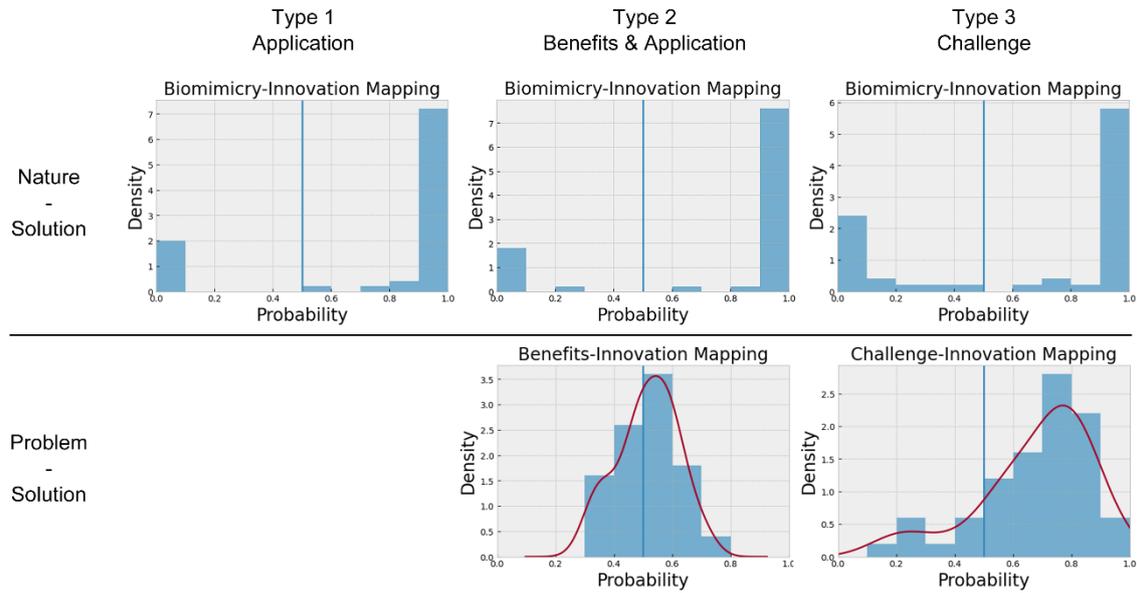

Figure 8: Probability distribution of the domains relevancy of the generated flying car concepts

Table 5: Percentage of generated flying car concepts passing the evaluations

|  | Type-1 (Application) | Type-2 (Benefits & Application) | Type-3 (Challenge) |
|---|---|---|---|
| **Nature-Solution** | 80% | 80% | 66% |
| **Problem-Solution** | N/A | 58% | 84% |
| **Overall** | 80% | 50% | 62% |

### 5.3 Team Evaluation

We evaluated the remaining GPT-generated BID concepts passing the relevancy evaluation and narrow down to a small number of candidates for embodiment design, detailed design, and physical prototyping. We chose two concepts generated by each type and asked the flying car design team to evaluate them in two dimensions of feasibility and novelty.

Table 6 lists the concepts for the team to evaluate. Two concepts from the Type-1 generator that passes the evaluation are also included. These can provide an intuitive sense to the readers on the GPT-generated concepts as well as the performance of our approach. The model is capable of learning how to retrieve biological information that shares attributional or relational similarity with the target, and then map it to the target to generate solutions. For instance, the second concept in Table 6 proposes to mimic the lightweight skeleton of pterosaurs to design the flying car body. Meanwhile, both biomimicry and innovation descriptions could be not concise enough and contain extra information that is irrelevant to the analogy of our interest.

Table 6: Concepts for team evaluation

| | **Type-1 Concepts (generated from Application as problem domain input)** |
|---|---|
| 1# | *Biomimicry:* Swimming animals must produce thrust to move forward. The thrust is generated by moving the tail fin, which works like a propeller.<br>*Innovation:* The flying car was inspired by the way a shark swims through water. The shark tail fin moves back and forth to create thrust, just like a propeller. However, the shark fin has shape-shifting skin inside that helps to smooth out the flow of water while swimming. The flying car has a similar shape-shifting skin inside its tail fin to help with thrust, similarly to the shark fin. It has a rectangular profile when at rest. When it moves through the air, the tail fin smoothly changes shape to a wedge-shape profile, similar to a shark fin. The two different shapes affect the way air moves over the tail fin, creating thrust. |
| 2# | *Biomimicry:* Pterosaurs are considered the dinosaurs of the Mesozoic era, because they soared as high as 50 m above the ground. They were the largest flying animals ever, with wingspans of approximately 3 m. Their wings were thin and long, and they had a bony breastbone that was shaped like an upside-down 'V'. Their forelimbs were short, but their claws were long and sharp, helping them to grab on to branches.<br>*Innovation:* The flying car has a body that is similar in shape to pterodactyls, with a body designed to control drag, lift, and thrust. It also has a lightweight hull and a propeller to generate thrust. The vehicle's hull is constructed of high-performance carbon fiber, inspired by the lightweight skeletons of pterosaurs. The propeller is mounted on a pivoting arm that is controlled by a joystick. The entire assembly weighs approximately 35 pounds and looks similar to a parasail. The propeller is 16 inches in diameter and is powered by a 930cc marine engine. |
| | **Type-2 Concepts (generated from Benefits and Application as problem domain input)** |
| 3# | *Biomimicry:* Hummingbirds are lightweight, small animals that need to stay aloft while carrying food to their nestlings. They do this by using the forces of lift and drag, which are affected by airspeed and wing shape. At low airspeeds, hummingbird wings provide thrust by moving back and forth. At high airspeeds, their curved outer wing surfaces trap air, creating low drag and enabling them to stay aloft at high speeds.<br>*Innovation:* The FLY Bot is made of carbon fiber and epoxy and weighs only 9.4 ounces (266 grams). It is powered by a brushless DC motor and has three wings that are made of a porous, lightweight material. The wings have a curved profile similar to a hummingbird, which allows it to have high lift at low speeds while also being able to move the robot quickly through the air. The high-aspect-ratio wings have porous materials in the inner cores, including carbon fiber, materials that are similar to the sponges found in sand dollars and the nano structure of sea sponges, which allows them to be lightweight as well as strong and flexible. The wings are also connected to close-fitting anchors that are used to control aerodynamic forces. This allows the robot to fly at high speeds of up to 27 miles per hour. Overall, the FLY-BOT weighs only a pound and its wingspan is 10 feet long. |
| 4# | *Biomimicry:* Many birds are able to fly for very long periods of time without tiring. This is because birds are able to keep their heart rates low and oxygen consumption at a minimum. Their lightweight and porous bones allow their bodies to keep warm, which helps to conserve energy.<br>*Innovation:* The lightweight carbon fiber body of the flying car has many of the same porous and lightweight bone characteristics found in a bird. It has a carbon fiber body with a porous, lightweight lattice of carbon fiber ribs that support the vehicle and minimize drag while in flight. The ribs are attached via carbon fiber spars that are held together by a unique adhesive that was inspired by gecko feet, which allow the vehicle to stay attached when flying through the air. |

| | **Type-3 Concepts (generated from Challenge as problem domain input)** |
|---|---|
| 5# | *Biomimicry:* Some insects, such as dragonflies and fruit flies, are known for their high maneuverability and high speed flight. This agility is enabled by a wing geometry that gives the wing a high aspect ratio (length to width ratio) and a thin structure.<br>*Innovation:* The wing design of the flying car is inspired by the wing shape of dragonflies and fruit flies. It is made of lightweight carbon fiber that has the same stiffness as steel, but is 60% lighter. The wing is able to effectively integrate air and ground characteristics into a single wing design, which could lead to more efficient use of materials, increased maneuverability, and more. The main wing is wedge-shaped and has a very small aspect ratio, similar to fruit flies and dragonflies. The small aspect ratio significantly reduces the total wing surface area, generating greater lift for a given wing area. The high lift generates significantly less drag and less induced drag, which allows the vehicle to achieve greater efficiency and a better aerodynamic performance. |
| 6# | *Biomimicry:* The flying fish stores energy for its leap above the water surface. It does this by bending its body in a "U" shape, with its tail lying flat. When it is ready to jump, it springs its tail upwards and propels itself through the air.<br>*Innovation:* The lightweight flying car has a horizontal propeller fitted with a folding propeller system, similar to a flying fish. The propeller folds when the vehicle is in driving mode, and unfolds when going into flying mode. When folded, the propeller is protected and the vehicle can travel over land and water. |

Moreover, we also added two samples of robotics innovations from AskNature as the benchmarking group for the team to evaluate. They are: "Safer aerial robot inspired by pufferfish"[8] and "Highly maneuverable drone inspired by swifts"[9]. This could make a better sense of how good our generated concepts are compared to the existing BID examples.

The rubrics for feasibility and novelty evaluation are shown in Table 7. For feasibility, we generally refer to the idea of workability [50] which represents the ability to implement the concept. On the other hand, novelty is the degree that measures how common the concept has already been seen in the target domain and its related industries. Note that by our metrics, a concept that had been expressed before but was never successfully implemented is still considered novel. Additionally, when evaluating the concepts, the team were asked to ignore the detailed numbers contained in the texts. This is because the PLMs are known to have limited capacity to perform arithmetic reasoning and could produce errors regarding the generated numbers [51, 52]. As we only want the team to focus on the analogies being retrieved and mapped in the concepts, the numbers can be disruptive and misleading for the evaluation process. The limitation of performing arithmetic logic will be further discussed in section 6.2.

Table 7: Feasibility and novelty rating rubrics

| Feasibility Rank | Description |
|---|---|
| 1 | The concept makes no sense from the engineering perspective. |
| 2 | The concept makes little sense with today's technology, but could be possible in the future. |
| 3 | The concept makes sense, but efforts are needed to work out a practical technical roadmap. |
| 4 | The concept makes good sense, and a technical roadmap can be easily established to realize it. |
| 5 | The concept makes perfect sense and there are existing tools, materials, or components to realize it. |

---

[8] https://asknature.org/innovation/safer-aerial-robot-inspired-by-pufferfish/
[9] https://asknature.org/innovation/highly-maneuverable-drone-inspired-by-swifts/

| Novelty Rank | Description |
|---|---|
| 1 | Solution exists and is commonly seen in the target domain (target domain = the target product described in the concept) |
| 2 | Solution exists but is uncommon in target domain. |
| 3 | New features are proposed for target domain, but similar approaches or technology can be commonly seen in related industries (related industries = flying car, drone, automobile, robotics, etc.) |
| 4 | New features are proposed for target domain, and similar approaches or technology can be rarely found in related industries. |
| 5 | New features are proposed, and no similar approaches or technology can be found nowadays. |

After the survey, 9 out of the 10 returned results are useable and are analyzed in Figure 9 by the generator types. The team gave relatively high novelty levels but only intermediate feasibility levels to the comparison group, which is reasonable because a lot of samples from AskNature are not directly workable with existing materials or components. Comparing to this, Type-1 and Type-2 concepts got lower feasibility score but higher novelty score, while the ones generated from Type-3 are considered more feasible but less novel. Type-1 and Type-2 generations are more open than Type-3, and this is similar to our finding from Figure 4 in Section 4.1.

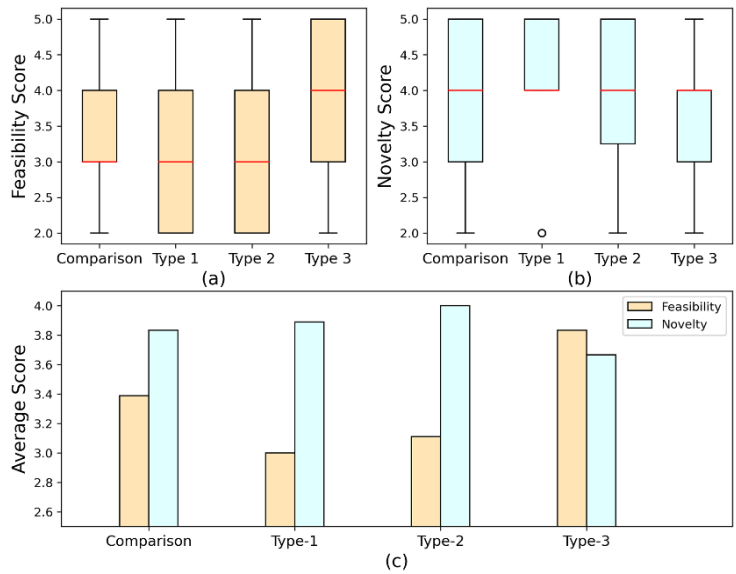

Figure 9: Human evaluation results: distribution of (a) feasibility scores; (b) novelty scores; (c) average score of feasibility and novelty

Overall, the results show that feasible and novel BID flying car concepts can be generated by fine-tuned GPT for the engineering team for selecting the one to move forward for embodiment and detailed design.

With the evaluation, we want to select a concept for further development. Concept 4# received the highest average novelty score among all concepts with an acceptable feasibility score. It suggests mimicking the porous structure of the bone of birds for the material design in the flying car and this is beyond the expertise of the team. Eventually concept 2# is selected to proceed for embodiment and detailed design. 2# draws analogy from the pterosaurs that were the largest known flying animal on earth and yet they can fly fast. The concept suggests mimicking the body shape, wing structure and lightweight skeleton of pterosaurs to support the design of a lightweight flying car. Additionally, while we were looking into pterosaurs to design the flying car, we noticed that the folding mechanism of the pterosaurs' skeletons is particularly inspiring. During unfolding the wings, the pterosaurs also spread its limbs because a part of the wings' membrane is attached to the limbs to increase the wing surface. The resulting concept not only mimics the wing structure to design flexible wings for the flying car, but also learns from the folding mechanism of the pterosaurs to combine wheels and propellers into the same set of propulsion system to further reduce weight. Interestingly, although we did not input any problem when generating 2# through the Type-1 generator, it still provides clues and inspirations for us to design a lightweight flying car. Figure 10 depicts the initial sketch drawn by the first author that illustrates the early embodiment design of the flying car concept and its folding process from the aerial mode to land mode of the flying car compared to that of the pterosaurs.

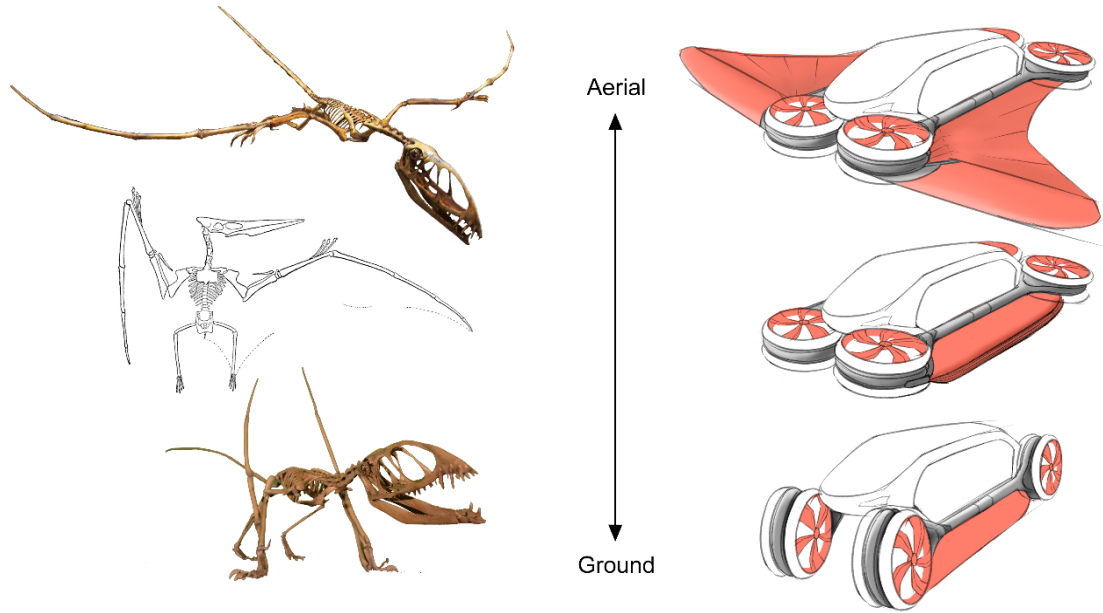

Figure 10: Design sketch based on concept 2#

This example showcases that human-AI collaboration in design concept generation does not necessarily stop at applying what the algorithm directly offers. Designers can take the knowledge and inspiration provided by AI to discover more inspirations and further develop the concept, add more details, and optimize it for their own project scenario.

## 6  Discussion

### 6.1 Implications of the Results

This study has demonstrated the method and workflow to employ GPT-3 for bio-inspired design process. Comparing to current BID approaches, the proposed workflow enables automatic retrieval and mapping of the biological source as design analogy. This work also contributes to the design community as an early study to utilize PLM in conceptual design, showing promising capability and potential of the cutting-edge AI technology for generating novel design concepts.

#### 6.1.1 Automatic Retrieval and Mapping of Biological Inspiration

Biomimicry as design inspiration is good to have but difficult to get. This is because most designers are not knowledgeable enough about the characteristics of the wide range of biological systems in nature. The gap between engineering and biology results in challenges that designers encounter [4, 5]:

the challenges of finding and recognizing relevant biological information, and of understanding the biological information in a way that is applicable to engineering design. Designers attempting to apply a BID approach usually have to manually look up a lot of literatures to find relevant biological knowledge. This search activity is inefficient because designers can only vaguely describe the information needed [5]. Moreover, experience and training are needed for designers to retrieve, understand, and transfer biological information like BID experts.

Prior studies of computational approaches supporting BID processes mostly focus on biological knowledge retrieval from specific assigned knowledge bases. For example, [34] extracts causally related function knowledge in a corpus of biological book, [37] developed a keyword pushing approach based on the corpus collected from two biological dictionaries. However, the knowledge retrieval approaches in these works do not support further interpretation and inference on how designers could understand and transfer the biological knowledge for the specific engineering problem of interest. Our NLP-based approach (described in section 3) can not only automate the retrieval of biological knowledge, but also the mapping from biological information to engineering design. This provides a clear insight of what biological systems and characteristics could be used as inspiration for a design problem of interest, and how they could potentially be transferred into engineering solutions. With the AskNature innovations as fine-tuning dataset, GPT-3 picks up the reasoning of biological knowledge retrieval and mapping, and helps designers who have little knowledge and experience of BID to think like a BID expert.

Moreover, the knowledge encoded in a specific biological corpus cannot cover all possible topics of biology [37], and thus limits the sources of biological analogies. In contrast, large language models like GPT-3 have already been pre-trained on a massive textual dataset involving knowledge from most possible domains and topics. The pre-training data of GPT-3 includes around 400 billion text tokens collected from web, internet-based books, and Wikipedia data [40], and the model can retrieve the encoded knowledge in the way requested. This means that besides the AskNature data for fine-tuning, the model can also leverage a large variety of biological and engineering information that can be found on the internet. After fine-tuning with the AskNature dataset, the model could retrieve the encoded knowledge of both biology and engineering and synthesize them in the way of BID. Although the paper only presented the case study of flying car design, the proposed approach can also be used for the interests of other engineering domains like medical equipment, materials, coating, and more. Table 8 illustrates several input design problems from other engineering fields than flying car and the generated results from our broader experiments of exploring the general applicability of our approach.

Table 8: Examples of concepts generated for more diverse engineering fields and problems

| Generator Type | Prompt | Example of Generated Solution |
| --- | --- | --- |
| Type-1 | *Application:* Implanted prosthetics | *Biomimicry:* The mammalian inner ear contains thousands of hair cells that serve as mechanical sensors, converting fluid flows within the inner ear into nerve signals that are sent to the brain.<br>*Innovation:* The bio-implantable synthetic ear is made of a hydrogel scaffold that contains three different types of mechanosensory cells: pillar cells, pillar/brush cells, and cilia. These cells use mechanoelectrical transduction to convert mechanical signals into electrical signals, similar to biological systems. This enables the device to detect a wide range of sound frequencies. |
| Type-2 | *Benefit:* Anti-fouling<br>*Application:* Marine coating | *Biomimicry:* The surface of lotus leaves has microscopic bumps that trap dirt and water and prevent it from sticking to the surface of the leaf. This is known as lotus effect, also known as dirt repellency. |
|  |  | *Innovation:* The coating mimics the lotus effect by embedding nanoparticles that have a hydrophobic end and a hydrophilic end. The hydrophobic end binds to the surface of the material and the hydrophilic end is able to attract water. As water is attracted, the nanoparticles repel oil and dirt, which helps to prevent biofouling. |
| Type-3 | *Challenge:* Thin-film and 3D microbatteries have been proposed as compact and efficient energy-storage solutions, but they often suffer from a poor rate performance and a short cycle life. | *Biomimicry:* Mussels are able to stick to wet surfaces such as rocks or ship hulls with their strong, 'stringy' byssus threads, despite the wet and viscous environment. This is largely attributed to the chitin-protein link, which forms high-performance adhesives that are fast and strong, and yet environmentally adaptive.<br>*Innovation:* The battery has a self-regulating chitin-protein adhesive between the electrode and electrolyte layers that is inspired by mussels. The adhesive can regulate the ionic conductivity between these layers, causing a more polycondensate-like structure that allows for the storage of hydrogen gas. This allows for the battery to have a high energy density and high discharge voltage without sacrificing cycle life. |

**6.1.2 PLM for Engineering Design**

The application of AI and natural language processing has been explored extensively in the design research community. A recent review [53] has shown that NLP techniques has been utilized to assist many processes in design, including the discovery and definition of user needs, the exploration of technical design opportunities, and the generation and evaluation of design concepts. Regenwetter et al. [54] reviewed design synthesis using deep generative models (e.g., Variational Autoencoder, Generative Adversarial Network), which generate shape or geometry that could be useful for the later design stages of embodiment design and detail design. Other studies also show a promising performance to utilize AI in design management [55, 56]. However, the adoption of large pre-trained language models and generative transformer-based models is still an underexplored topic for design research [53, 54, 57]. This paper aims to fill in this gap by employing generative pre-trained transformer for bio-inspired design, while our recent study has explored the applications of GPT-2 and GPT-3 to generate new design concepts using a product design data repository [58, 59].

One advantage of large PLMs like GPT-3 is that they can understand and leverage the reasoning within the linguistic task of interest, with only a small number of the task examples. In engineering design, a large amount of high-quality data structured for a certain domain and design task is seldomly available, thus limiting the application of most generative models today [54]. Large PLMs, on the other hand, is already capable of understanding and generating human language in general through pre-training and need only a small amount of task samples for fine-tuning or few-shot learning to generate the content of interest. This can be helpful for teaching the model about different design heuristics of divergent thinking, such as biomimicry, for generating design concepts [58].

Moreover, thanks to the extraordinary size of the pre-training corpus, a large variety of knowledge can be leveraged for design tasks. This does not limit to the knowledge related to the design of interest but also the remote knowledge that has rarely occurred together with the design target. This advantage enables PLMs to contribute to divergent thinking during conceptual design stage by performing a wide range of design-by-analogy tasks, including BID that was introduced in this paper. Formerly, the exploration of far-field analogy sources can be supported by knowledge graphs or semantic networks [19-21, 53]. However, such tools can only help the retrieval of knowledge as semantic stimuli and still require designers to perform the mapping and generate concepts. PLMs could resolve the problem by synthesizing a large variety of knowledge encoded in the pre-training corpus in the analogical reasoning learned through fine-tuning or few-shot learning.

**6.2 Limitations**

Although this paper has shown good performance of employing GPT-3 for BID, such application of large language models in design is still in an early stage. This work can be preliminary and limited in the following aspects:

1. *The lack of technicality in the knowledge.* Although the pre-training corpus of GPT-3 ensures the variety and richness of both biological and engineering knowledge, they are mostly common-sense knowledge but not from technical reports or publications. Future works could explore the pre-training of language models with more technical knowledge of biology and engineering.
2. *Computationally very expensive.* The model of GPT-3 is very large which requires the computing power beyond personal affordability to train or run the model. Currently the model can only be retrieved, fine-tuned, and employed through OpenAI API.
3. *Varied quality and feasibility of the results.* The generation with GPT-3 picks up a highly explorative strategy and some of the results may not make much sense from an engineering perspective. Moreover, large language models have been shown to struggle to perform arithmetic tasks and produce errors in logic [51, 52]. In this case, this limitation comes in the form of producing infeasible numbers from the engineering perspective. Although researchers have found a way to improve it by training verifiers [52], it requires a focused training with datasets of arithmetic problem-solving examples. Nevertheless, in the divergence process of BID concept generation, it is essential to explore inspirational analogies that designers could hardly think of without extensive reserves of biological knowledge. Designers are free to explore, verify, adjust, and test the generated concepts from engineering perspective, as we have demonstrated in section 5.3.
4. *Biases in the generation.* The fine-tuned GPT-3 may pick up biases regarding what biomimicry sources to retrieve. The biological systems that have historically been used as biomimicry sources to solve a similar problem (especially those occurred in the AskNature database) will have a higher chance to be retrieved. Increasing the temperature parameter during generation could mitigate the problem by encouraging more diverse results, but could also result in more vulnerable quality of the generated concept descriptions.
5. *Comparison of different models and strategies.* A few hundreds of samples are not enough to fine-tune other popular language models (e.g., BERT and GPT-2). Thus, we are unable to test and compare different PLMs in their performances of generating or evaluating BID concepts. Moreover, we only introduced three model fine-tuning and concept generation strategies in this paper. There could be more and more flexible strategies for fine-tuning GPT-3 for concept generation. For example, solution-based approach that uses biological systems as a starting point could be explored.

In addition, the case study is limited in its scale and scope. The evaluation of 10 engineers from one design project is insufficient to provide statistical significance. As the next step of this research, we plan to engage more and more diverse engineers to evaluate the concepts generated by the fine-tuned GPT for different design problems and scenarios. These will allow us to develop more systematic understandings on the enabling or conditioning factors on the performance of our methodology and shed light on additional insights for improving the methodology and guiding its uses in context.

Therefore, these limitations of our work present new research opportunities and future research directions. In general, we hope our exploratory work presented here can inspire more research on the use or development of generative AI for biologically inspired design, and design-by-analogy in general.

# 7   Concluding Remarks

This paper has introduced a data-driven AI methodology for bio-inspired design concept generation. The core of the methodology is to fine-tune the generative pre-trained transformers for different concept generation situations with different data inputs and use them to automatically generate design concepts in natural language texts. We demonstrated that the PLMs could be customized to retrieve the knowledge and reasoning for biologically inspired design process, bridging the gap between engineering and biology. The case study shows reasonable performance of the methodology in generating good quality BID concepts in intelligible natural language.

Our work responds to the recent calls in the design research community to adopt pre-trained language models for design research and applications [53, 54, 57]. To the best of our knowledge, this study is the first to employ generative pre-trained transformers for bio-inspired design. In the time that transformer-based foundation models [60] start to reshape both academia and industry, we believe it is timely and crucial to explore the opportunities that it offers to engineering design research. By showing evidence of great potentials, the authors hope this paper to inspire more studies into this area and data-driven innovation [61].